\documentclass[11pt]{article}
\usepackage[margin=1in]{geometry}
\usepackage{amsmath,amssymb,amsthm}
\usepackage[round,authoryear]{natbib}
\usepackage{booktabs}
\usepackage{graphicx}
\usepackage{microtype}
\usepackage[hidelinks]{hyperref}
\usepackage{xcolor}
\usepackage{caption}
\newtheorem{conjecture}{Conjecture}

\newcommand{\NE}{\mathrm{NE}}
\newcommand{\KL}{\mathrm{KL}}
\newcommand{\Hent}{\mathcal{H}}
\newcommand{\Rnad}{R\text{-}NaD}
\newcommand{\Vstar}{V^\star}

\title{\textbf{The Curvature Shadow: An Apparent Failure of Maximum-Entropy\\ Equilibrium Selection is a Removable Artifact}\\[0.7em]
{\large\itshape A diagnostic study of the Kuhn poker gap in Regularized Nash Dynamics,\\ separating entropy shortfall from landscape flatness}}
\author{Luis Leal\\[0.2em]\texttt{wichofer89@gmail.com}}
\date{}

\begin{document}
\maketitle
\vspace{-1.5em}

\begin{abstract}
\noindent
In two-player zero-sum games whose Nash equilibria form a convex set, regularized solvers such as Regularized Nash Dynamics ($\Rnad$) empirically select the maximum-entropy member---the information projection (I-projection) of a uniform reference onto the Nash set. On a panel of small games this match is exact, with one apparent exception: in Kuhn poker $\Rnad$ lands at bluff coordinate $0.180$ while the maximum-entropy member sits at $0.201$, a coordinate gap of $\approx\!0.021$, even though $\Rnad$ attains $99.7\%$ of the maximum entropy. We ask whether this gap is a genuine selection bias or an artifact, and answer it quantitatively. We show that for selection on a one-dimensional Nash manifold the coordinate gap factorizes as $\mathrm{gap}\approx\sqrt{2\delta/\kappa}$, where $\delta$ is the solver's entropy shortfall and $\kappa$ is the curvature of the entropy landscape at its peak. Across five games this relation holds to within $2\times10^{-4}$ ($<1\%$ relative error). The four matrix games have $\delta\approx0$ ($\Rnad$ reaches the maximum-entropy member exactly) and therefore no gap regardless of curvature; only the sequential game (Kuhn) has $\delta>0$. A causal sweep of the magnet strength drives $\delta\to0$ and the gap toward zero along the predicted curve (fitted gap-vs.-$\delta$ scaling exponent $0.50$, $R^2>0.999999$, against the exact H-flat prediction of $1/2$), until the dynamics destabilize at a stability floor---behavior consistent with a \emph{removable shortfall} and inconsistent with a fixed bias. We quantify the curvature half of the law from measured curvatures and flag a moving-target pitfall in the natural Tsallis-entropy experiment. The Kuhn gap is thus the curvature-shadow of a small, removable entropy shortfall on an unusually flat peak; the I-projection account is upheld up to a flatness-limited residual.
\end{abstract}

\section{Introduction}
A line of recent work studies \emph{which} Nash equilibrium a learning algorithm converges to when the equilibrium set of a two-player zero-sum (2p0s) game is not a single point but a convex polytope, all of whose members share the game value $\Vstar$. A robust empirical finding, established against analytic ground truth and across solver families by \citet{whichnash2026}, is that regularized methods in the family of magnetic mirror descent~\citep{sokota2023} and Regularized Nash Dynamics ($\Rnad$)~\citep{perolat2021,perolat2022} select the \emph{maximum-entropy} equilibrium: the member that maximizes Shannon entropy over the polytope, equivalently the information projection (I-projection) of a uniform reference onto the Nash set. On a panel of tabular games this prediction is met exactly---with one apparent exception. In Kuhn poker, $\Rnad$ converges to a bluff frequency of $0.180$, whereas the maximum-entropy equilibrium has bluff frequency $0.201$; the resulting coordinate gap of $\approx\!0.021$ persists even though the converged profile attains $99.7\%$ of the maximum attainable entropy and is an \emph{exact} Nash equilibrium (zero exploitability).

This note diagnoses that gap. The question is whether $\Rnad$ selects a point that is \emph{systematically displaced} from the maximum-entropy member (a genuine bias in the selection target), or whether the gap is an \emph{artifact} of two innocuous quantities: a small residual entropy shortfall, and the local flatness of the entropy landscape that converts that shortfall into a visible coordinate offset. We formalize this as a competition between two hypotheses:
\begin{itemize}
\item[\textbf{H-flat:}] $\Rnad$ computes the maximum-entropy member up to a small \emph{removable} entropy shortfall $\delta$; the coordinate gap equals $\sqrt{2\delta/\kappa}$ and vanishes as $\delta\to0$.
\item[\textbf{H-bias:}] $\Rnad$ selects a member systematically off the maximum-entropy point; the gap persists at a nonzero value even as $\delta\to0$.
\end{itemize}

\paragraph{Contributions.}
(i) We derive an exact local decomposition of the selection gap on a one-dimensional Nash manifold, $\mathrm{gap}\approx\sqrt{2\delta/\kappa}$ (Section~\ref{sec:decomp}). (ii) We verify it to within $2\times10^{-4}$ across five games, and show that the four matrix games have $\delta\approx0$ (hence no gap), while only the sequential game has $\delta>0$ (Section~\ref{sec:gaplaw}). (iii) We provide a causal test on the shortfall: weakening the magnet drives $\delta\to0$ and the gap down the predicted curve, with a fitted gap-vs.-$\delta$ scaling exponent of $0.50$ over $14$ stable sweep points (H-flat predicts exactly $1/2$; a fixed bias would drive it toward $0$), supporting H-flat, with the only obstruction a dynamical stability floor (Section~\ref{sec:eta}). (iv) We quantify the curvature half of the law from measured curvatures, and correct a natural Tsallis-entropy experiment whose target moves with the entropy index (Section~\ref{sec:curv}). The result is a defensible account of the apparent counterexample, and a falsifiable prediction for sequential games with sharply curved entropy peaks.

The study is deliberately narrow and diagnostic. It does not establish the I-projection claim in general (that remains a conjecture, framed below), nor does it identify the mechanism that produces the shortfall in sequential games; both are discussed as open problems.

\section{Preliminaries}
\paragraph{Games and the Nash set.}
We consider finite 2p0s extensive-form games with perfect recall. A behavioral strategy profile is $\sigma=(\sigma_I)_{I}$, one distribution per information set $I$. By the minimax theorem all Nash equilibria share a common value $\Vstar$; the set of equilibria $\NE$ is convex (a polytope), and in each game studied here it is a one-dimensional segment. We summarize a segment by a scalar \emph{selection coordinate} $c$, an affine readout of $\sigma$ chosen per game (e.g.\ the Kuhn bluff frequency); the segment is the curve $c\mapsto\sigma(c)$.

\paragraph{Exploitability.}
For player $i$ let $b_i(\sigma)$ be the value of $i$'s best response to $\sigma_{-i}$. We measure distance to equilibrium by $\textsc{NashConv}(\sigma)=b_0(\sigma)+b_1(\sigma)$, which is $0$ exactly at a Nash equilibrium and positive otherwise.

\paragraph{Entropy and the I-projection.}
Write the per-profile mean entropy
\begin{equation}
H(\sigma)\;=\;\frac{1}{|\mathcal{I}|}\sum_{I}\Hent(\sigma_I),
\qquad \Hent(p)=-\textstyle\sum_a p_a\log p_a .
\end{equation}
The I-projection of the uniform reference $u$ onto $\NE$ is $\arg\min_{\sigma\in\NE}\sum_I \KL(\sigma_I\,\|\,u_I)$. Since $\KL(\sigma_I\|u_I)=\log n_I-\Hent(\sigma_I)$ with $n_I=|{\rm actions}(I)|$, and $\sum_I\log n_I$ is constant on $\NE$, this projection coincides exactly with the maximum-(total, hence mean)-entropy member of $\NE$~\citep{csiszar1975}. We denote its coordinate $c^\star=\arg\max_c H(\sigma(c))$ and its entropy $H^\star=H(\sigma(c^\star))$.

\paragraph{Regularized Nash Dynamics.}
$\Rnad$---the regularized-dynamics framework introduced by \citet{perolat2021}, named and scaled to Stratego by \citet{perolat2022}---is magnetic mirror descent~\citep{sokota2023} with a \emph{moving} reference. With learning rate $\mathrm{lr}$, magnet strength $\eta$, and $c_\eta=1/(1+\mathrm{lr}\cdot\eta)$, the per-infoset update is
\begin{equation}
\sigma_{t+1,I}\;\propto\;\exp\!\Big(c_\eta\log\sigma_{t,I}+c_\eta\,\mathrm{lr}\,\hat q_{t,I}+(1-c_\eta)\log\rho_{I}\Big),
\label{eq:rnad}
\end{equation}
where $\hat q_{t,I}$ is the vector of counterfactual action values~\citep{zinkevich2007} at $I$ normalized by the infoset reach probability, and $\rho$ is the reference policy. For a \emph{fixed} reference $\rho$ the iteration converges to the quantal response equilibrium (QRE)~\citep{mckelvey1995} of the entropy-regularized game; periodically resetting $\rho$ to the current policy (every \texttt{ref\_period} iterations) yields a sequence of QREs converging to a Nash equilibrium. The uniform reference biases selection toward high-entropy members; the I-projection account is the conjecture that this bias selects exactly the maximum-entropy member.

\begin{conjecture}[I-projection selection; Conjecture~1 of \citealp{whichnash2026}]\label{conj:iproj}
$\Rnad$ initialized and referenced at the uniform strategy selects the I-projection of the uniform reference onto $\NE$, i.e.\ the maximum-entropy equilibrium.
\end{conjecture}

\section{The curvature decomposition}\label{sec:decomp}
Restrict $H$ to the Nash segment and expand around its maximum. Writing $\kappa=-\,\tfrac{d^2}{dc^2}H(\sigma(c))\big|_{c^\star}>0$ for the \emph{peak curvature},
\begin{equation}
H(\sigma(c))\;=\;H^\star-\tfrac12\kappa\,(c-c^\star)^2+O\!\big((c-c^\star)^3\big).
\end{equation}
A point on the segment at entropy shortfall $\delta:=H^\star-H(\sigma(c))$ therefore satisfies $\tfrac12\kappa(c-c^\star)^2\approx\delta$, i.e.
\begin{equation}
\boxed{\;\mathrm{gap}:=|c-c^\star|\;\approx\;\sqrt{2\delta/\kappa}\;.}
\label{eq:law}
\end{equation}
The expansion is valid because the maximum is \emph{interior}: in all five games $c^\star$ lies strictly inside the coordinate range (Table~\ref{tab:games}), where every action probability that varies with $c$ is strictly positive, so $H(\sigma(c))$ is smooth at $c^\star$ and the first-order term $\tfrac{d}{dc}H(\sigma(c))\big|_{c^\star}$ vanishes by stationarity. Were $c^\star$ instead pinned to a boundary of the segment, the leading behavior of the shortfall would be linear and the gap would scale as $\delta/|H'(c^\star)|$ rather than $\sqrt{2\delta/\kappa}$; no game in our panel is in this regime.
Two observations make this more than an identity. First, \eqref{eq:law} applies to a point \emph{on} the Nash segment; a solver that converges to an exact equilibrium ($\textsc{NashConv}\!=\!0$) lands on the segment, so verifying \eqref{eq:law} for $\Rnad$'s output empirically confirms that its only deviation from $c^\star$ is the entropy shortfall---there is no transverse or off-manifold component to the gap. Second, \eqref{eq:law} cleanly separates the two ingredients of a gap: the \emph{shortfall} $\delta$ (why a solver stops short of maximum entropy) and the \emph{curvature} $\kappa$ (how strongly a fixed shortfall is amplified into a coordinate offset). H-flat predicts the gap is governed entirely by \eqref{eq:law} and vanishes with $\delta$; H-bias predicts a residual $|c-c^\star|$ that survives $\delta\to0$.

\section{Experimental setup}\label{sec:setup}
\paragraph{Engine.}
All computations use a tabular extensive-form engine that evaluates exact counterfactual values and full reach probabilities with no sampling and no function approximation (double-precision \texttt{numpy}). $\textsc{NashConv}$ is computed by exact best response. The engine and the notebook that produces every number and figure below are released with this paper.

\paragraph{Games and Nash families.}
We use five games whose Nash set is a one-dimensional segment (Table~\ref{tab:games}; full definitions in Appendix~\ref{app:games}): the sequential, imperfect-information game \texttt{kuhn} (Kuhn poker~\citep{kuhn1950}; $12$ information sets), and four single-infoset-per-player matrix games (\texttt{asym\_safe}, \texttt{pennies\_safe}, \texttt{two\_safe}, \texttt{dup\_action}) with analytically known segments. For each game the segment is given as an explicit parameterization $c\mapsto\sigma(c)$; we verify that every sampled member is an exact equilibrium ($\max_c \textsc{NashConv}(\sigma(c))\le1.1\times10^{-16}$).

\paragraph{Maximum-entropy coordinate and curvature estimation.}
The maximum-entropy coordinate $c^\star$ is computed exactly: in closed form for the matrix games (\texttt{pennies\_safe}: $c^\star=1/3$; \texttt{two\_safe}: $c^\star=1/2$; \texttt{dup\_action}: $c^\star=1/4$; \texttt{asym\_safe}: $c^\star=1/(3+4^{1/3})\approx0.2180$, from the stationarity condition $(1-3p)^3=4p^3$), and by high-precision golden-section maximization of the analytic family for \texttt{kuhn}, with $H^\star=H(\sigma(c^\star))$. For the curvature we evaluate $H(\sigma(c))$ on an $801$-point uniform grid over the coordinate range and estimate $\kappa$ by a least-squares quadratic fit to $H$ on the window $|c-c^\star|\le0.10\,(\text{range})$. (Taking $c^\star$ as the grid argmax instead would quantize it to the grid spacing and manufacture spurious $\sim\!10^{-4}$-level gaps in the matrix games.)

\paragraph{$\Rnad$ configuration.}
Unless stated otherwise we run $\Rnad$ from the uniform strategy and uniform initial reference with $\mathrm{lr}=1$, $\eta=0.5$, $\texttt{ref\_period}=60$, for $1.5\times10^4$ iterations ($2\times10^4$ in the $\eta$-sweep). The reported strategy is the final (last-iterate) policy, which $\Rnad$ converges to directly (no averaging).

\begin{table}[t]\centering\small
\caption{The five games and the verified curvature of their entropy landscapes. ``Nash-check'' is $\max_c\textsc{NashConv}(\sigma(c))$ over the parameterized family; $c^\star$ is the maximum-entropy coordinate; $\kappa$ is the peak curvature. Curvatures span an order of magnitude; notably, two matrix games are \emph{flatter} than Kuhn.}
\label{tab:games}
\begin{tabular}{llcccc}
\toprule
game & type & Nash-check & $c^\star$ & $H^\star$ & $\kappa$\\
\midrule
\texttt{kuhn}         & sequential & $1.1\times10^{-16}$ & $0.2008$ & $0.2614$ & $4.01$\\
\texttt{asym\_safe}   & matrix     & $1.1\times10^{-16}$ & $0.2180$ & $0.8489$ & $20.08$\\
\texttt{pennies\_safe}& matrix     & $0$                  & $0.3333$ & $0.8959$ & $2.27$\\
\texttt{two\_safe}    & matrix     & $0$                  & $0.5000$ & $1.0397$ & $2.01$\\
\texttt{dup\_action}  & matrix     & $0$                  & $0.2500$ & $0.8664$ & $4.02$\\
\bottomrule
\end{tabular}
\end{table}

\section{Results}
\subsection{Entropy landscapes and curvature}\label{sec:landscapes}
\begin{figure}[t]\centering
\includegraphics[width=\textwidth]{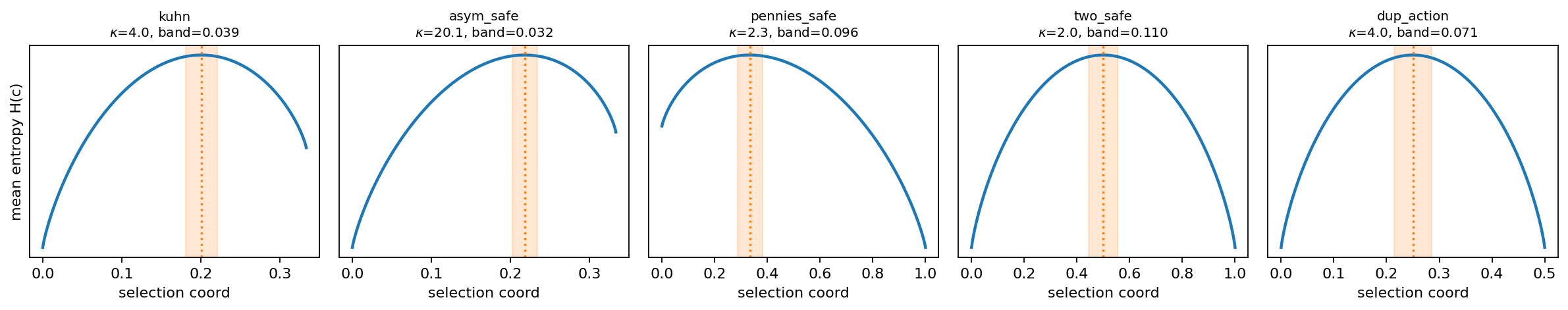}
\caption{Mean-entropy landscapes $H(\sigma(c))$ over each Nash segment. Dotted line: the maximum-entropy coordinate $c^\star$. Shaded: the band of coordinates within $99.7\%$ of $H^\star$. Flatter peaks (smaller $\kappa$) yield wider bands; Kuhn ($\kappa\!=\!4.0$) is flatter than the sharp matrix games but not the flattest overall.}
\label{fig:landscapes}
\end{figure}

Table~\ref{tab:games} reports the peak curvature of each game. The curvatures span an order of magnitude, from $\kappa\approx2$ (\texttt{two\_safe}, \texttt{pennies\_safe}) to $\kappa\approx20$ (\texttt{asym\_safe}). Crucially, Kuhn's peak ($\kappa\approx4$) is \emph{not} the flattest; two matrix games are flatter.\footnote{$\kappa$ carries the units of the selection coordinate and is therefore not invariant to reparameterizing the segment: under an affine change $c\mapsto ac$, $\kappa\mapsto\kappa/a^{2}$ while $\mathrm{gap}\mapsto a\cdot\mathrm{gap}$, so both sides of \eqref{eq:law} rescale consistently and the \emph{within-game} law is parameterization-free. Cross-game comparisons of $\kappa$ (such as ``flatter than Kuhn''), however, are relative to the fixed natural per-game coordinates of Appendix~\ref{app:games}; under range-normalized coordinates ($\kappa\cdot\mathrm{range}^2$), for instance, Kuhn's peak would itself be the flattest. The substantive claim of Section~\ref{sec:gaplaw}---that the matrix games have $\delta\approx0$ and hence no gap regardless of curvature---does not depend on the parameterization.} Figure~\ref{fig:landscapes} plots $H(\sigma(c))$ for each game and shades the band of coordinates within $99.7\%$ of $H^\star$. The band is wide where the peak is flat and narrow where it is sharp: a fixed entropy tolerance translates into a large coordinate spread exactly when $\kappa$ is small. This is the geometric content of \eqref{eq:law}---but, as the next subsection shows, flatness alone produces no gap.

\subsection{The gap is the curvature-shadow of an entropy shortfall}\label{sec:gaplaw}
\begin{figure}[t]\centering
\includegraphics[width=0.52\textwidth]{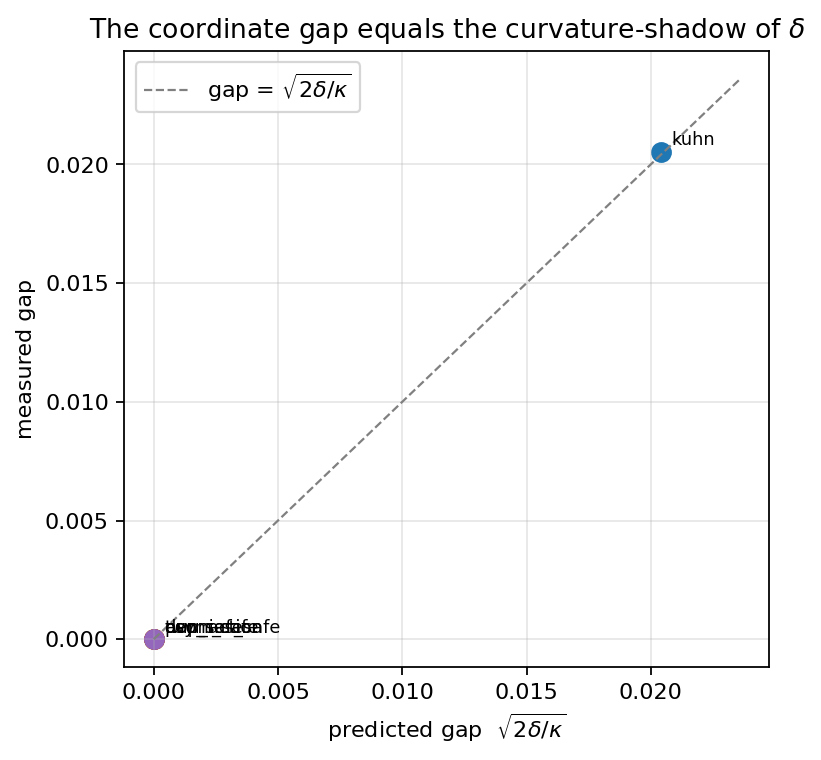}
\caption{Measured coordinate gap vs.\ the prediction $\sqrt{2\delta/\kappa}$ across the five games. Points lie on the diagonal; the matrix games cluster at the origin ($\delta\approx0$), Kuhn alone has a nonzero gap fully accounted for by its shortfall and curvature.}
\label{fig:gaplaw}
\end{figure}

We run $\Rnad$ on each game and measure its landed coordinate, entropy shortfall $\delta$, and coordinate gap, then compare the gap to the prediction $\sqrt{2\delta/\kappa}$ (Table~\ref{tab:gaplaw}, Figure~\ref{fig:gaplaw}). Two facts stand out.

First, the four \textbf{matrix games have $\delta\approx0$}: $\Rnad$ reaches the maximum-entropy member exactly, so there is no gap \emph{regardless of curvature}---including for \texttt{pennies\_safe} and \texttt{two\_safe}, whose peaks are flatter than Kuhn's. Flatness is therefore necessary but not sufficient for a gap; it amplifies a shortfall but cannot create one.

Second, \textbf{Kuhn alone has $\delta>0$} ($\delta=8.3\times10^{-4}$), and there the measured gap ($0.0205$) agrees with $\sqrt{2\delta/\kappa}$ ($0.0204$) to within $2\times10^{-4}$ ($<1\%$ relative; the small excess is the expected third-order asymmetry of the peak). The prediction is also robust to the one methodological choice in it: varying the curvature-fit window from $\pm5\%$ to $\pm20\%$ of the coordinate range moves $\kappa$ only across $3.99$--$4.09$ ($2.4\%$ spread), a predicted-gap spread of $0.0202$--$0.0204$; since every computation here is deterministic, this window sensitivity is the relevant error bar on the prediction, and the measured gap is consistent with the law within it. The gap is thus not an independent selection bias: it is the coordinate image of a small entropy shortfall under the local curvature. Two checks close the loop. The shortfall is a property of the regularized fixed point rather than finite training: at fixed $\eta=0.5$, both $\delta$ and the landed coordinate are unchanged to five significant figures from $5\times10^{3}$ to $8\times10^{4}$ iterations---only changing $\eta$ (Section~\ref{sec:eta}) moves them. And the on-manifold premise of \eqref{eq:law} holds directly: the converged profile coincides with the analytic family member at its landed coordinate to within $\sim\!10^{-11}$ per action probability, so the entire deviation from $c^\star$ lies along the segment. That the shortfall appears only in the single sequential, imperfect-information game points to the reach-weighted regularization of the sequential setting as its origin---a hypothesis we return to in Section~\ref{sec:discussion}.

\begin{table}[t]\centering\small
\caption{$\Rnad$ selection per game and the gap law. All runs reach a numerically exact equilibrium (\textsc{NashConv}$\le3\times10^{-12}$). The coordinate gap equals $\sqrt{2\delta/\kappa}$ throughout; matrix games have $\delta\approx0$ and no gap, Kuhn has $\delta>0$.}
\label{tab:gaplaw}
\begin{tabular}{lcccccc}
\toprule
game & coord & $c^\star$ & $\delta$ & $\kappa$ & gap & $\sqrt{2\delta/\kappa}$\\
\midrule
\texttt{kuhn}          & $0.1802$ & $0.2008$ & $8.3\times10^{-4}$ & $4.0$  & $0.0205$ & $0.0204$\\
\texttt{asym\_safe}    & $0.2180$ & $0.2180$ & $\approx0$         & $20.1$ & $0.0000$ & $0.0000$\\
\texttt{pennies\_safe} & $0.3333$ & $0.3333$ & $\approx0$         & $2.3$  & $0.0000$ & $0.0000$\\
\texttt{two\_safe}     & $0.5000$ & $0.5000$ & $\approx0$         & $2.0$  & $0.0000$ & $0.0000$\\
\texttt{dup\_action}   & $0.2500$ & $0.2500$ & $\approx0$         & $4.0$  & $0.0000$ & $0.0000$\\
\bottomrule
\end{tabular}
\end{table}

\subsection{Causal test on the shortfall: weakening the magnet}\label{sec:eta}
\begin{figure}[t]\centering
\includegraphics[width=0.6\textwidth]{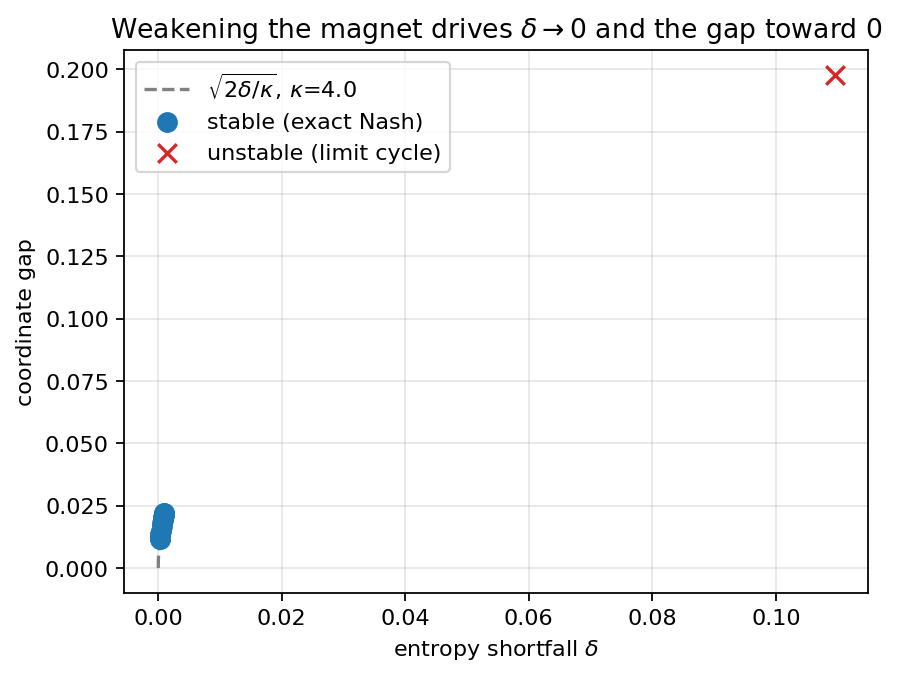}
\caption{Kuhn magnet sweep in the $(\delta,\mathrm{gap})$ plane. Dashed: the prediction $\sqrt{2\delta/\kappa}$ for $\kappa=4.0$. Stable runs (blue) lie on the curve and move toward the origin as $\eta$ decreases; the unstable run (red $\times$) leaves the curve once the dynamics limit-cycle.}
\label{fig:eta}
\end{figure}

H-flat predicts the gap is \emph{removable}: drive $\delta\to0$ and the gap should follow $\sqrt{2\delta/\kappa}\to0$. We sweep $15$ magnet strengths $\eta\in[0.10,0.80]$ on Kuhn (Table~\ref{tab:eta}, Figure~\ref{fig:eta}). As $\eta$ falls from $0.80$ to $0.18$, the shortfall shrinks monotonically from $9.8\times10^{-4}$ to $2.6\times10^{-4}$ and the gap from $0.0222$ to $0.0115$, with every point a numerically exact equilibrium (\textsc{NashConv}$\le3\times10^{-12}$) and lying on the $\sqrt{2\delta/\kappa}$ curve for the measured $\kappa=4.0$. At the last stable point, $\eta=0.15$, $\delta$ and the gap tick up marginally ($2.7\times10^{-4}$, $0.0117$)---a precursor of the instability---while still lying exactly on the curve. Below that the regularization is too weak to stabilize the dynamics: at $\eta=0.10$ the iterate limit-cycles ($\textsc{NashConv}=0.11$) and the coordinate collapses to the segment boundary. This is the documented bias-versus-stability frontier of fixed/slow magnets.

\paragraph{Functional-form test.}
The sweep permits a sharper, quantitative discrimination between the hypotheses than trajectory inspection. H-flat predicts $\mathrm{gap}=\sqrt{2/\kappa}\,\delta^{1/2}$, i.e.\ a log--log slope of exactly $1/2$ in the $(\delta,\mathrm{gap})$ plane; under H-bias the gap tends to a nonzero constant as $\delta\to0$, driving the local slope toward $0$. A least-squares fit of $\log\mathrm{gap}$ on $\log\delta$ over the $14$ stable sweep points yields a scaling exponent of $0.5013$ with $R^2>0.999999$, and the fit's intercept independently recovers $\kappa=3.88$, within $3\%$ of the quadratic-fit value $4.01$. Because the pipeline is deterministic (exact tabular dynamics, no sampling), this functional-form test---rather than seed-based inferential statistics, which would be vacuous here---is the appropriate statistical assessment, and it comes out decisively for H-flat.

The trajectory is exactly what H-flat predicts and H-bias forbids: the gap does not converge to a nonzero residual but decreases along the predicted curve toward zero as the shortfall is driven out, and the only obstruction to reaching zero is a property of the \emph{dynamics} (the stability floor), not of the selection target. Within the stable regime, $\Rnad$'s selected point approaches the maximum-entropy member as $\eta\to0$, consistent with Conjecture~\ref{conj:iproj}.

\begin{table}[t]\centering\small
\caption{Magnet sweep on Kuhn ($\kappa=4.0$), $15$ values of $\eta$. Reducing $\eta$ reduces the shortfall and the gap along $\sqrt{2\delta/\kappa}$ while the iterate remains a numerically exact equilibrium; $\delta$ falls monotonically down to $\eta=0.18$ and ticks up marginally at the last stable point $\eta=0.15$, a precursor of the stability floor.}
\label{tab:eta}
\begin{tabular}{lccccc}
\toprule
$\eta$ & coord & $\delta$ & gap & \textsc{NashConv} & stable\\
\midrule
$0.80$ & $0.1785$ & $9.8\times10^{-4}$ & $0.0222$ & $0.0000$ & yes\\
$0.70$ & $0.1788$ & $9.5\times10^{-4}$ & $0.0219$ & $0.0000$ & yes\\
$0.60$ & $0.1794$ & $9.1\times10^{-4}$ & $0.0214$ & $0.0000$ & yes\\
$0.55$ & $0.1797$ & $8.7\times10^{-4}$ & $0.0210$ & $0.0000$ & yes\\
$0.50$ & $0.1802$ & $8.3\times10^{-4}$ & $0.0205$ & $0.0000$ & yes\\
$0.45$ & $0.1809$ & $7.8\times10^{-4}$ & $0.0199$ & $0.0000$ & yes\\
$0.40$ & $0.1817$ & $7.2\times10^{-4}$ & $0.0191$ & $0.0000$ & yes\\
$0.35$ & $0.1827$ & $6.4\times10^{-4}$ & $0.0180$ & $0.0000$ & yes\\
$0.30$ & $0.1841$ & $5.5\times10^{-4}$ & $0.0167$ & $0.0000$ & yes\\
$0.25$ & $0.1859$ & $4.4\times10^{-4}$ & $0.0149$ & $0.0000$ & yes\\
$0.22$ & $0.1873$ & $3.6\times10^{-4}$ & $0.0135$ & $0.0000$ & yes\\
$0.20$ & $0.1883$ & $3.1\times10^{-4}$ & $0.0125$ & $0.0000$ & yes\\
$0.18$ & $0.1893$ & $2.6\times10^{-4}$ & $0.0115$ & $0.0000$ & yes\\
$0.15$ & $0.1891$ & $2.7\times10^{-4}$ & $0.0117$ & $0.0000$ & yes\\
$0.10$ & $0.0031$ & $1.1\times10^{-1}$ & $0.1977$ & $0.1131$ & \textbf{no} (limit cycle)\\
\bottomrule
\end{tabular}
\end{table}

\subsection{The curvature half, quantified, and the Tsallis caveat}\label{sec:curv}
\begin{figure}[t]\centering
\begin{minipage}{0.49\textwidth}\centering
\includegraphics[width=\textwidth]{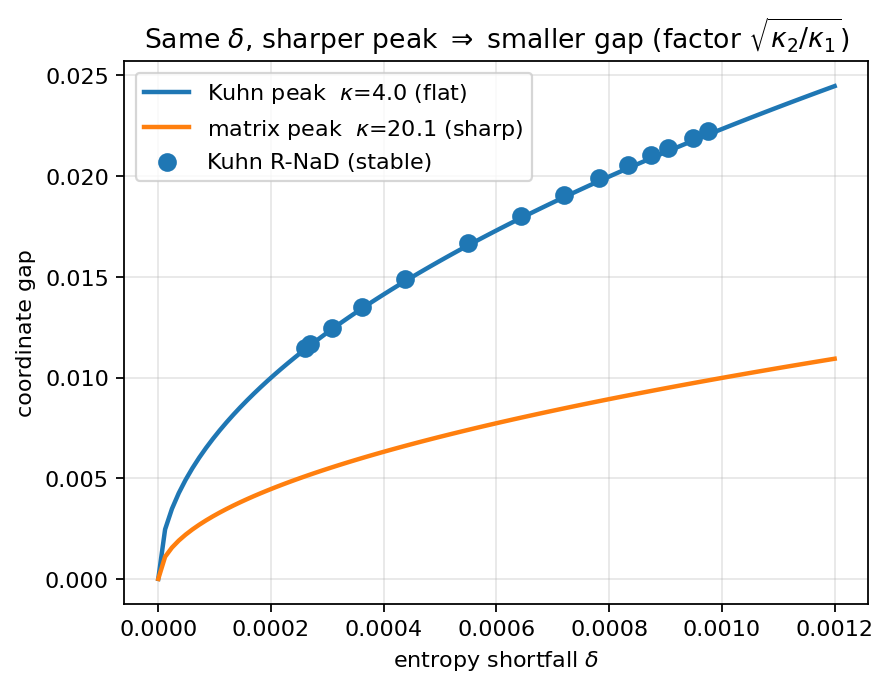}
\caption{Curvature amplification. The same shortfall $\delta$ maps to a smaller gap on a sharper peak; Kuhn's stable $\eta$-sweep points lie on the $\kappa\!=\!4.0$ curve. A $\kappa\!=\!20.1$ peak would give a $\approx2.24\times$ smaller gap for the same $\delta$.}
\label{fig:curv}
\end{minipage}\hfill
\begin{minipage}{0.49\textwidth}\centering
\includegraphics[width=\textwidth]{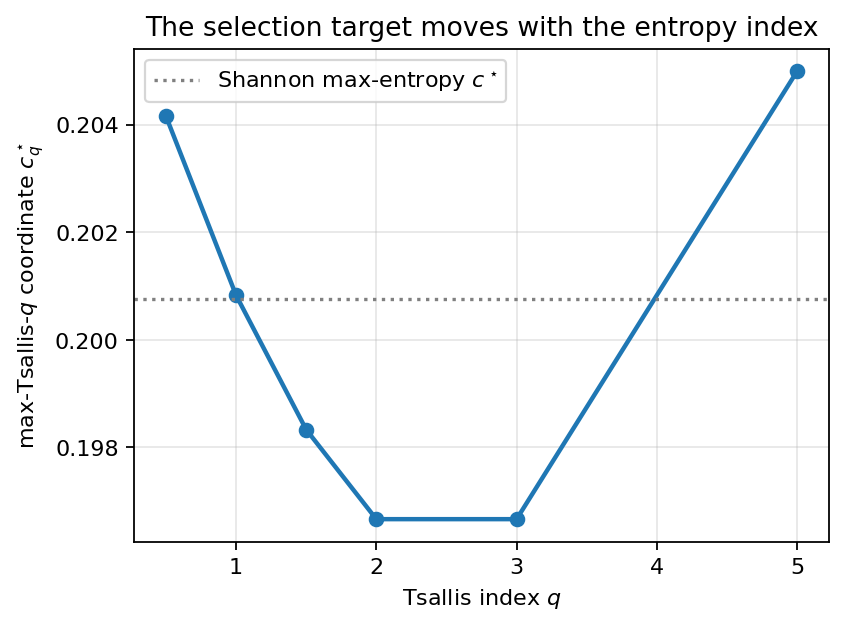}
\caption{The selection target moves with the entropy index. Maximum-Tsallis-$q$ coordinate $c^\star_q$ over the fixed Kuhn family vs.\ $q$; dotted line is the Shannon target. A Tsallis-regularized solver must be compared to $c^\star_q$, not the Shannon $c^\star$.}
\label{fig:tsallis}
\end{minipage}
\end{figure}

The $\eta$-knob varies $\delta$ at fixed curvature. The complementary direction is $\kappa$ itself. We were unable to construct a reliable \emph{sequential} curvature dial---the natural candidate, Kuhn with varying bet size, does not yield a stable one-dimensional family (for larger bets the bluffing family collapses toward a unique point), and the matrix games, which do span a range of $\kappa$, have $\delta\approx0$ and hence no gap to amplify. We therefore quantify the curvature dependence directly from \eqref{eq:law} and the measured curvatures. Figure~\ref{fig:curv} overlays the $\sqrt{2\delta/\kappa}$ curves for Kuhn's peak ($\kappa=4.0$) and a sharp matrix-game peak ($\kappa=20.1$): a shortfall identical to Kuhn's, on the sharp peak, would yield a gap $\sqrt{\kappa_{\text{sharp}}/\kappa_{\text{Kuhn}}}\approx2.24\times$ smaller. Kuhn's moderately flat peak is, quantitatively, why its (small) shortfall is visible as a coordinate gap at all.

\paragraph{The Tsallis moving-target caveat.}
A tempting way to vary $\kappa$ directly is to regularize $\Rnad$ with a Tsallis-$q$ entropy in place of Shannon, sharpening the peak by tuning the index $q$. This is unsound unless done carefully: changing the regularizer changes the \emph{fixed point itself}. A Tsallis-$q$-regularized solver targets the maximum-Tsallis-$q$ member, whose coordinate $c^\star_q$ \emph{moves with $q$}. Figure~\ref{fig:tsallis} shows $c^\star_q$ over the fixed Kuhn family ranging across $0.197$--$0.205$ as $q$ varies in $[0.5,5]$. Comparing such a solver against the \emph{Shannon} maximum-entropy coordinate would conflate this moving target with the curvature effect and produce an uninterpretable trend; any Tsallis curvature dial must compare each run to its own matched target $c^\star_q$.

\section{Discussion}\label{sec:discussion}
\paragraph{The gap is an artifact, not a bias.}
Across all five games the coordinate gap equals $\sqrt{2\delta/\kappa}$; the matrix games have $\delta\approx0$ and no gap even when their peaks are flatter than Kuhn's, and Kuhn's gap is fully accounted for by its shortfall and curvature. The causal $\eta$-sweep shows the gap is removable: it tracks the predicted curve toward zero as $\delta\to0$, with no fixed residual and a fitted gap-vs.-$\delta$ scaling exponent of $0.50$ ($R^2>0.999999$) against the exact H-flat prediction of $1/2$. Together these support H-flat and are inconsistent with H-bias. With respect to Conjecture~\ref{conj:iproj}, the I-projection account is upheld up to a flatness-limited residual: $\Rnad$'s selected point converges to the maximum-entropy member as the magnet weakens, and the apparent Kuhn ``counterexample'' is the visible shadow of a sub-$10^{-3}$ entropy shortfall on a peak flat enough to magnify it into a $0.02$ coordinate offset.

\paragraph{Scope of the curvature relation.}
We emphasize that $\mathrm{gap}\approx\sqrt{2\delta/\kappa}$ is a \emph{local geometric identity}---a second-order Taylor readout of how a scalar entropy shortfall $\delta$ projects onto a coordinate offset on a single manifold of peak curvature $\kappa$---and not an empirical claim that any offset-\emph{generating} process scales cleanly with $\kappa$ across games. The two are distinct: the identity relates $\delta$ and $\kappa$ on a fixed peak, whereas how strongly the underlying dynamics displace the selected point can depend on many game-specific factors and need not order by $\kappa$ across a panel. We make only the former, narrower claim here; questions about the cross-game curvature dependence of the displacement itself are outside this diagnostic and, where later studied, are not expected to follow a clean monotone law.

\paragraph{Why does the shortfall appear only in the sequential game?}
The matrix games (one information set per player, reach probability $1$) reach the maximum-entropy member exactly; only Kuhn, with reach-weighted counterfactual updates across $12$ information sets, leaves a residual $\delta>0$. We \emph{hypothesize} that the reach-weighting in \eqref{eq:rnad}---which scales the regularization felt at each infoset by how often it is reached---distorts the moving-reference QRE sequence away from the unweighted maximum-(mean-)entropy member, leaving a small shortfall that vanishes only in the $\eta\to0$ limit. This is a hypothesis, not a result: we have not isolated the mechanism, and establishing it (and its dependence on tree depth and reach skew) is the natural next step.

\section{Limitations}\label{sec:limitations}
\begin{itemize}
\item \textbf{Curvature half is quantified, not directly swept.} We did not run a causal $\kappa$-sweep on a sequential game, because no reliable sequential family with a tunable sharp peak \emph{and} a nonzero shortfall was available; Kuhn bet-size variants proved degenerate. The curvature dependence is therefore established via \eqref{eq:law} and measured curvatures, not a second solver-driven dial.
\item \textbf{Single sequential exemplar.} The phenomenon ($\delta>0$) appears in one game (Kuhn). Whether other sequential games exhibit it, and how $\delta$ scales with depth and reach skew, is untested.
\item \textbf{Mechanism unproven.} The reach-weighting explanation for $\delta>0$ is a hypothesis.
\item \textbf{Objective-dependence of the target.} The maximum-entropy member depends on the entropy used (Shannon vs.\ Tsallis vs.\ reach-weighted entropy); $c^\star$ shifts accordingly (Figure~\ref{fig:tsallis}). Our analysis fixes the unweighted mean Shannon entropy as the objective.
\item \textbf{Stability floor.} The $\eta$-sweep cannot reach $\delta=0$ because slow/fixed magnets destabilize below $\eta\approx0.15$; the extrapolation to a zero gap is therefore a limit statement, not a directly observed endpoint. Empirically, the degradation announces itself: $\delta$ ticks up marginally at the last stable point before the dynamics limit-cycle.
\item \textbf{Selection is initialization-dependent.} All results concern $\Rnad$ initialized and referenced at the uniform strategy, as in Conjecture~\ref{conj:iproj}. Under random initial policies the solver still converges to numerically exact equilibria ($\textsc{NashConv}\le4\times10^{-12}$) but lands across the segment (coordinates $0.057$--$0.258$ over ten random initializations)---consistent with the initialization/reference being the selection mechanism itself, but a reminder that the I-projection account is a statement about the uniform regime, not about $\Rnad$ from arbitrary starts. Since the dynamics are otherwise deterministic (exact tabular updates, no sampling), this initialization sensitivity is the meaningful analogue of a multi-seed study; seed-based error bars would be vacuous here.
\item \textbf{Tabular only.} All results are exact-tabular; whether the decomposition survives sampling and function approximation is open.
\item \textbf{Local curvature estimate.} $\kappa$ is a least-squares quadratic fit over a fixed window; the $2\times10^{-4}$-level agreement in Table~\ref{tab:gaplaw} indicates the quadratic approximation is adequate for the observed small shortfalls, but $\kappa$ is a local quantity.
\end{itemize}

\section{Conclusion}
The apparent failure of maximum-entropy equilibrium selection in Kuhn poker is not a selection bias but the curvature-shadow of a small, removable entropy shortfall. On a one-dimensional Nash manifold the selection gap factorizes exactly as $\sqrt{2\delta/\kappa}$; matrix games have no shortfall and hence no gap, while Kuhn's sub-$10^{-3}$ shortfall, amplified by a moderately flat entropy peak, produces the observed $0.02$ offset, which a magnet sweep drives toward zero. The I-projection account of regularized selection survives the apparent counterexample.

\noindent\textbf{Falsifiable prediction.} On a sequential game with a sharply curved entropy maximum over its Nash segment, $\Rnad$ with a uniform reference should exhibit a coordinate match to the maximum-entropy member that is tighter than Kuhn's by the factor $\sqrt{\kappa/\kappa_{\text{Kuhn}}}$, approaching exactness as $\kappa$ grows or as the magnet weakens. Constructing such a game---or a correctly matched-target Tsallis-regularized solver---would convert the curvature half of the law from a quantified prediction into a directly observed causal sweep.

\paragraph{Reproducibility.}
Every number, table, and figure is produced by a single self-contained notebook (\texttt{curvature\_test.ipynb}) over the embedded tabular engine; hyperparameters are as in Section~\ref{sec:setup} and game definitions are in Appendix~\ref{app:games}.

\appendix
\section{Game and family definitions}\label{app:games}
All matrix games specify the row player's (P0, maximizer) payoff matrix $M$; the column player (P1, minimizer) sees nothing. The selection coordinate $c$ and the explicit Nash family $\sigma(c)$ are listed; each family was verified to have $\textsc{NashConv}\le1.1\times10^{-16}$ on a $40$-point grid.

\paragraph{\texttt{kuhn} (sequential).} Standard Kuhn poker: deck $\{$J$,$Q$,$K$\}$, one card each, ante $1$, bet size $1$. The player-0 Nash family is the one-parameter family with bluff $\eta=\Pr(\text{bet}\mid\text{Jack})\in[0,1/3]$, value-bet $\Pr(\text{bet}\mid\text{King})=3\eta$, and call frequency $\Pr(\text{call}\mid\text{Queen, facing bet})=1/3+\eta$, with the remaining infosets pinned and player~1's strategy unique; the coordinate is $c=\eta$. Value $\Vstar=-1/18$.

\paragraph{\texttt{asym\_safe} (matrix).} $M=\big[[3,-1],[-1,1],[\tfrac13,\tfrac13]\big]$. Rows r0,r1 form a skewed matching-pennies forcing P1 to play $(\tfrac13,\tfrac23)$; r2 is a value-matching safe row. Family $\sigma(c)=\big(\,c,\,2c,\,1-3c\,\big)$ for P0 with P1 $=(\tfrac13,\tfrac23)$; coordinate $c=\Pr(\text{r0})\in[0,\tfrac13]$. (\emph{Asymmetric} segment: $c^\star\neq$ uniform.)

\paragraph{\texttt{pennies\_safe} (matrix).} $M=\big[[1,-1],[-1,1],[0,0]\big]$. Matching pennies (r0,r1) plus a safe row r2. Family $\sigma(c)=\big(\tfrac{1-c}{2},\tfrac{1-c}{2},c\big)$ for P0, P1 $=(\tfrac12,\tfrac12)$; coordinate $c=\Pr(\text{r2})\in[0,1]$.

\paragraph{\texttt{two\_safe} (matrix).} $M=\big[[1,-1],[-1,1],[0,0],[0,0]\big]$. Matching pennies plus two safe rows. Family $\sigma(c)=\big(\tfrac{1-c}{2},\tfrac{1-c}{2},\tfrac{c}{2},\tfrac{c}{2}\big)$ for P0, P1 $=(\tfrac12,\tfrac12)$; coordinate $c=$ total safe mass $\in[0,1]$.

\paragraph{\texttt{dup\_action} (matrix).} $M=\big[[1,1,-1],[-1,-1,1]\big]$. Columns c0,c1 are duplicates, giving P1 a one-parameter family. Family $\sigma(c)=\big(c,\tfrac12-c,\tfrac12\big)$ for P1, P0 $=(\tfrac12,\tfrac12)$; coordinate $c=\Pr(\text{c0})\in[0,\tfrac12]$.

\section{The R-NaD update and metrics}\label{app:rnad}
The per-infoset update is \eqref{eq:rnad} with $c_\eta=1/(1+\mathrm{lr}\cdot\eta)$, counterfactual action values $\hat q_{t,I}$ normalized by infoset reach, and reference $\rho$ reset to the current policy every \texttt{ref\_period} iterations (a fixed $\rho$ would converge to the QRE; the moving reset yields the Nash-converging QRE sequence). $\textsc{NashConv}(\sigma)=b_0(\sigma)+b_1(\sigma)$ with $b_i$ the exact best-response value for player $i$; it is $0$ at equilibrium. Mean entropy is the unweighted average of per-infoset Shannon entropies. All quantities are computed in exact tabular form without sampling.


\begin{thebibliography}{9}
\small
\bibitem[Csisz\'ar(1975)]{csiszar1975} I.~Csisz\'ar. I-divergence geometry of probability distributions and minimization problems. \textit{The Annals of Probability}, 3(1):146--158, 1975. \href{https://doi.org/10.1214/aop/1176996454}{doi:10.1214/aop/1176996454}
\bibitem[Kuhn(1950)]{kuhn1950} H.~W. Kuhn. A simplified two-person poker. In \textit{Contributions to the Theory of Games}, vol.~1, pp.~97--103. Princeton University Press, 1950. \href{https://doi.org/10.1515/9781400881727-010}{doi:10.1515/9781400881727-010}
\bibitem[Leal(2026)]{whichnash2026} L.~Leal. Which Nash equilibrium? Solver-dependent selection on zero-sum Nash polytopes. \textit{arXiv:2606.28308}, 2026.
\bibitem[McKelvey \& Palfrey(1995)]{mckelvey1995} R.~D. McKelvey and T.~R. Palfrey. Quantal response equilibria for normal form games. \textit{Games and Economic Behavior}, 10(1):6--38, 1995. \href{https://doi.org/10.1006/game.1995.1023}{doi:10.1006/game.1995.1023}
\bibitem[Perolat et al.(2021)]{perolat2021} J.~Perolat, R.~Munos, J.-B.~Lespiau, S.~Omidshafiei, M.~Rowland, P.~Ortega, N.~Burch, T.~Anthony, D.~Balduzzi, B.~De~Vylder, G.~Piliouras, M.~Lanctot, and K.~Tuyls. From Poincar\'e recurrence to convergence in imperfect information games: Finding equilibrium via regularization. In \textit{ICML}, pp.~8525--8535, 2021.
\bibitem[Perolat et al.(2022)]{perolat2022} J.~Perolat, B.~De~Vylder, D.~Hennes, E.~Tarassov, F.~Strub, et al. Mastering the game of Stratego with model-free multiagent reinforcement learning. \textit{Science}, 378(6623):990--996, 2022. \href{https://doi.org/10.1126/science.add4679}{doi:10.1126/science.add4679}
\bibitem[Sokota et al.(2023)]{sokota2023} S.~Sokota, R.~D'Orazio, J.~Z. Kolter, N.~Loizou, M.~Lanctot, I.~Mitliagkas, N.~Brown, and C.~Kroer. A unified approach to reinforcement learning, quantal response equilibria, and two-player zero-sum games. In \textit{ICLR}, 2023. \href{https://doi.org/10.48550/arxiv.2206.05825}{doi:10.48550/arxiv.2206.05825}
\bibitem[Zinkevich et al.(2007)]{zinkevich2007} M.~Zinkevich, M.~Johanson, M.~Bowling, and C.~Piccione. Regret minimization in games with incomplete information. In \textit{NeurIPS}, 2007.
\end{thebibliography}
\end{document}